# Diagnostic Posture Control System for Seated-Style Echocardiography Robot

Yuuki Shida[1*], Masami Sugawara[1], Ryosuke Tsumura[2], Haruaki Chiba[3], Tokuhisa Uejima[4]

Hiroyasu Iwata[5]

1 Graduate School of Creative Science and Engineering, Waseda University, Tokyo 169-8050, Japan

2 Global Robot Academia Laboratory, Waseda University, Tokyo 169-8050, Japan

3 NSK Ltd Technology Development Department 1New Field Products Development Center Technology Development Division Headquarters, Kanagawa 251-8501, Japan

4 The Cardiovascular Institute, Tokyo, 106-0031, Japan

5 Faculty of Science and Engineering, Waseda University, Tokyo 169-8050, Japan

* Corresponding author: Yuuki Shida (yuuki-shida@iwata.mech.waseda.ac.jp)

## Abstract

**Purpose** Conventional robotic ultrasound systems were utilized with patients in supine positions. Meanwhile, the limitation of the systems is that it is difficult to evacuate the patients in case of emergency (e.g., patient discomfort and system failure) because the patients are restricted between the robot system and bed. Then, it is ideal that the patient undergoes the examination in the sitting position in terms of safety. Therefore, we validated a feasibility study of seated-style echocardiography using a robot.

**Method** Preliminary experiments were conducted to verify the following two points: (1) the possibility of obtaining cardiac disease features in the sitting posture as well as in the conventional examination, and (2) the relationship between posture angle and physical burden. For reducing the physical burden, two unique mechanisms were incorporated into the system: (1) a leg pendulum base mechanism to reduce the load on the legs when the lateral bending angle increases, and (2) a roll angle division by a lumbar lateral bending and thoracic rotation mechanisms.

**Results** Preliminary results demonstrated that adjusting the diagnostic posture angle enabled us to obtain the views, including cardiac disease features, as in the conventional examination. The results showed that the body burden increased as the posture's lateral bending angle increased. The results also demonstrated that the body load reduction mechanism incorporated in the results could reduce the physical load in the seated echocardiography.

**Conclusion** These results showed the potential of the seated-style echocardiography robot.

# 1. Introduction

Heart disease has become the most common disease globally in terms of deaths. According to a World Health Organization (WHO) [1] survey, 17.9 million people died of heart disease in 2019. The mortality rate of heart disease can be significantly improved by early detection and treatment. In response to this situation, transthoracic echocardiography often referred to as cardiac ultrasound, has become the modality of choice in the initial assessment of cardiac disease because it is noninvasive, easy to use, and provides high-resolution imaging and real-time feedback [2]. However, ultrasonography, including transthoracic echocardiography, is highly challenging because of its sophisticated procedure. As a result, physicians and sonographers must have advanced experience and knowledge.

To solve the problems mentioned above, a wide range of robot-assisted technologies for ultrasound examinations have been developed. These robot systems have focused on several applications such as the carotid artery, liver, fetal echocardiography, cardiac tamponade, and other generic sites [3]-[13]. However, most of these robot systems are designed to be used with the patient in the left lateral decubitus or supine position, which is the recommended position for conventional examination methods. Those configurations were mainly applied through a serial robotic manipulator or gantry-style (Fig 1 (a)). In the case that the robot performs the examination in that position, the patient is positioned between the robot and the bed. Therefore, given that the patient's posture is restricted, emergency evacuation in case of patient discomfort or robot failure is difficult, which is insufficient for safety during the examination. Therefore, it is ideal that the patient undergoes the examination in a sitting position rather than in the left lateral decubitus or supine positions in terms of safety and emergency evacuation because the patient can immediately leave the robot system (Fig 1 (b)). In the conventional examination, the patient needs to be in the supine position first, and the examination is performed in the left lateral decubitus position when diagnostic images cannot be obtained clearly since the heart is hindered by other organs such as the lungs. By examining in the left lateral supine position, the heart is slightly moved so that it is not hindered by other organs. This is achieved by adjusting the direction of gravity applied to the heart. We hypothesize that the same phenomenon can be produced by adjusting the angle of the patient's posture, even in the sitting posture. If the quality of diagnostic images is ensured in the sitting posture, the safety of the robot system for assisting the ultrasonography can be guaranteed.

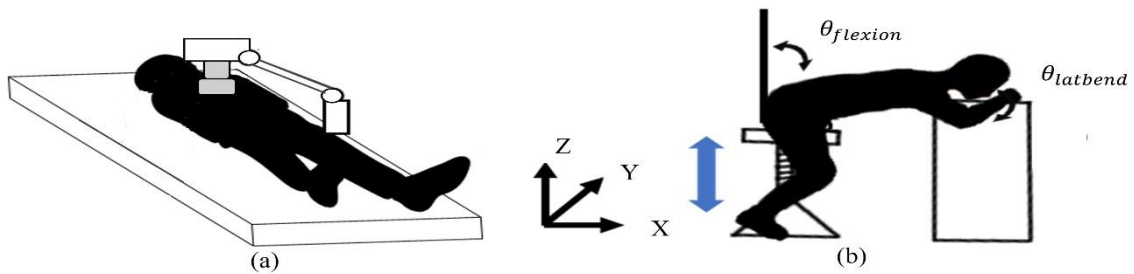

**Fig 1** Echocardiography robot and subject placement (a) supine (b) seated

Therefore, the purpose of this study is to establish the proof-of-concept of the seated-style echocardiography robot system. We experimentally analyzed the diagnosable posture angle at which diagnostic images visualizing the features of cardiac diseases can be acquired with healthy subjects. Based on the analysis, we proposed a seated diagnostic posture control system that enables the adjustment of the optimal posture angle in the sitting position, thereby visualizing the features of cardiac disease and decreasing the physical load occurring in the sitting position.

## 2. Analysis of diagnosable posture angles

### 2.1. Experimental analysis for diagnosable posture angle

The purpose of this experimental analysis is to verify whether it is possible to acquire ultrasound images that can extract the features of cardiac diseases in the sitting posture and to ensure ultrasound images with diagnostic qualities. According to the clinical expert in the field of Echocardiogram, to diagnose myocardial infarction, valvular disease, and cardiomegaly, which are the major diseases observable by echocardiography, it is necessary to observe three features: 1) ventricular motion, 2) dynamics and shape of the valve, and 3) enlargement of the ventricular wall. Then, we first acquired the parasternal long-axis and apical four-chamber views, which are the basic diagnostic views in which those three features are depicted. After the acquisition, we assessed that those three features are portrayed in those basic two views when the posture angle changes. With six healthy subjects, those views were acquired by a clinical expert in the field of echocardiography. Each view was acquired in the left lateral decubitus position and sitting posture based on ten conditions. A medical ultrasound system (EPIQ7, Philips, Netherland) and a matrix array sector probe (X5-1, Philips, Netherland) were used for the ultrasound image acquisition. An optical tracking sensor (Trio V120, OptiTrack, Japan) attached to the ultrasound probe was used to measure the position and angle of the ultrasound probe. A diagnostic posture angle adjustment table was used to change the angle of the sitting posture. The detailed flow of the experiment is described below.

(1) The subject is placed in the left lateral decubitus position. Then we acquire images in the parasternal long-axis view and the apical 4-chamber view.

(2) The subject is placed in the sitting posture and on the diagnostic posture angle adjustment table. As shown in Fig 2(a), we performed the following process with probe tracking to obtain the actual patient's posture angle: i) with the ensiform process as the starting point of the probe scanning, the probe was moved left and right and up and down to the nipple position; ii) planar approximation is applied to the tracked probe positions, and then the actual posture angle of the subject was calculated. In this study, the X-axis is defined as the front-back direction of the body when standing vertically, and the y-axis is defined as the left-right direction of the body, as shown in Fig 1(b). Note that $\theta_{roll}$ and $\theta_{pitch}$ are defined as shown in Fig. 2(b).

(3) The probe is moved to the position where the parasternal long-axis view and the apical four-chamber view can be acquired. Then, the subject stops breathing, and ultrasound images are acquired for two seconds while the subject is holding his breath. Each view is acquired three times.

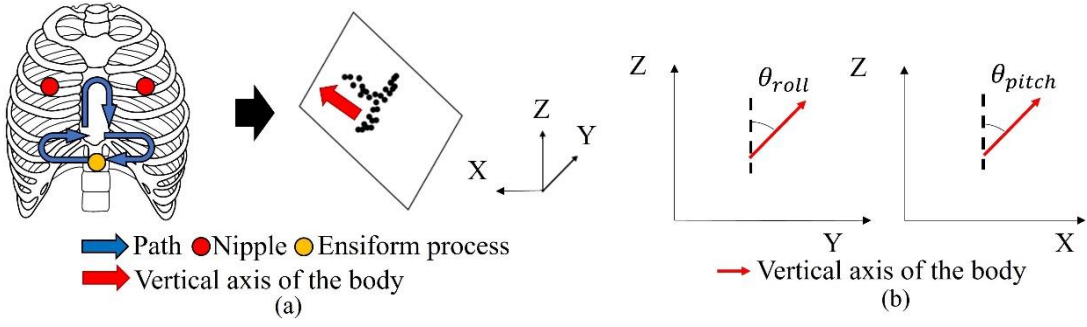

**Fig 2** Marker tracking calibration (a) Probe travel path and plane approximation (b) Calculation of $\theta_{\text{roll}}$, $\theta_{\text{pitch}}$

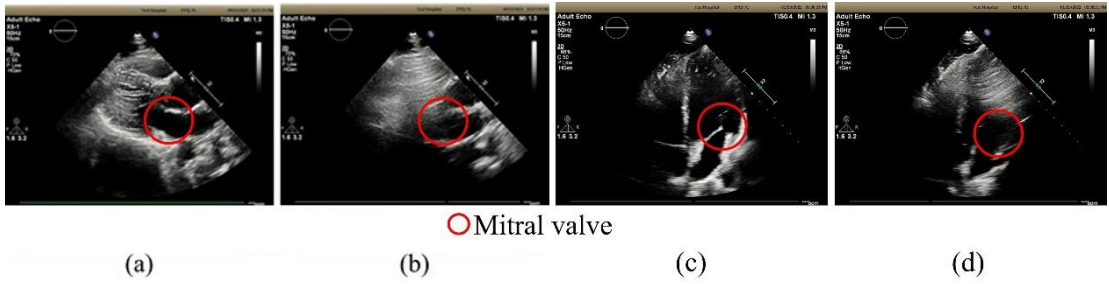

**Fig 3** Echo images in sitting posture; (a) Clear parasternal long-axis view; (b) Unclear parasternal long-axis view; (c) Clear apical four-chamber view; (d) Unclear apical four-chamber view

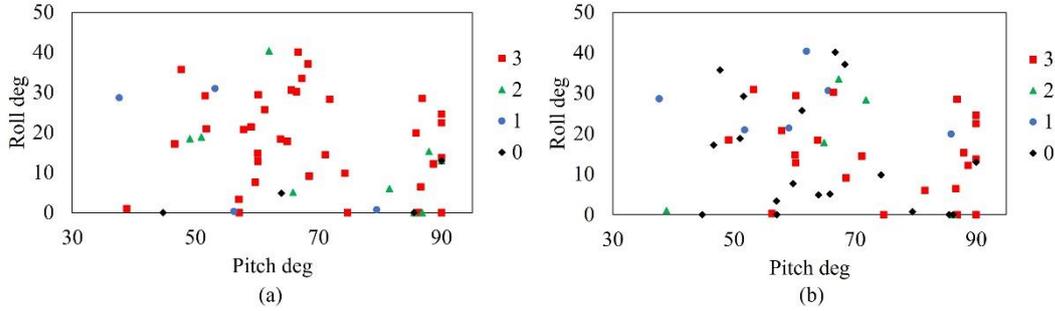

**Fig 4** Relationship between the number of diagnosable features as well as postur(a) The number of diagnosable features and posture in the parasternal long-axis view; (b) The number of diagnosable features and posture in the apical four-chamber view.

(4) The subject's sitting posture is changed in all nine conditions of the flexion angle $\theta_{\text{flexion}}$ (approx. 30, 60, 90°) in addition to the lateral bending angle $\theta_{\text{latbend}}$ (approx. 0, 15, 30°), as shown in Fig 1. We follow the same procedure given in (1) to obtain the parasternal long-axis and apical four-chamber views.

(5) The physician is evaluated on a three-point scale (diagnostic superiority, undecidable, and nondiagnostic quality) to determine whether the three aforementioned features can be visualized in each view. Based on this evaluation, the number of disease features that can be visualized is used to identify the sitting posture to ensure image clarity. Fig 3 presents representative ultrasound images with diagnostic superiority and nondiagnostic quality acquired in the sitting posture.

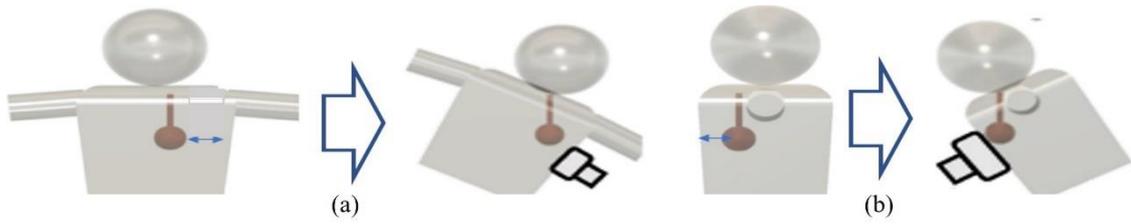

**Fig 5** Heart movement during body posture change; (a) roll; (b) pitch

Fig 4(a) and (b) show the relationship between the number of diagnosable features and posture in the parasternal long-axis and apical four-chamber views, respectively. From these results, it can be seen that it is possible to obtain each view where the three features can be detected even in the sitting posture. In the parasternal long-axis view, it was suggested that all the target disease features could be depicted in a posture with a roll angle of 10° to 30° and a pitch angle of 50° to 80°. On the other hand, in the apical four-chamber view, it was suggested that all the features of the target disease could be depicted in a posture with $\theta_{roll}$ = 10–20° and $\theta_{pitch}$ = 60–70°. As shown in Figs. 5 (a) and (b), when the roll/pitch angle is added to the body posture, the distance between the heart position and the probe becomes closer, preventing echo attenuation by the lungs. This is thought to make it easier to acquire a clear image. On the other hand, as the roll/pitch angle was increased, the number of diagnosable features of the heart decreased because the heart made more contact with the thoracic wall by increasing the posture angle than when the patient was in the left lateral supine position, which is the usual medical practice. This could have made the heart challenging to move and observe, which was observed in the apical four-chamber view rather than the parasternal long-axis view. Consequently, the apex of the heart observed in the apical four-chamber view is closer to the thoracic wall than any other part of the heart, and the postural angle causes the apex to contact the thoracic wall more quickly, affecting the motion of the heart. In this study, we analyzed the data of six subjects, and each of them had a different diagnosable posture angle, possibly due to the individual differences in heart position. According to medical knowledge [14], heart position varies among individuals and is classified into three types: right-leaning, center, and left-leaning. A left-leaning heart is closer to the thoracic wall than a right-leaning one. Therefore, the shift of the heart position due to posture variation is reduced. This difference is thought to cause individual differences in postural angles that can be diagnosed.

**2.2. Seated test body load verification**

In Sec. 2-1, we verified the diagnostic posture angles from the viewpoint of disease visualization. Conversely, some posture angles may impose a physical burden on the patient, making it difficult to perform the examination for a long time. In this experiment, the relationship between flexion and lateral flexion angles in the sitting posture of patients and their physical burden is verified by subjective evaluation using VAS. Based on the verification results, we derive the design requirements for a seated diagnostic posture control system that enables patients to undergo examination without any burden.

In this experiment, we used a support mechanism (Fig 6(a)) to adjust the flexion and the lateral

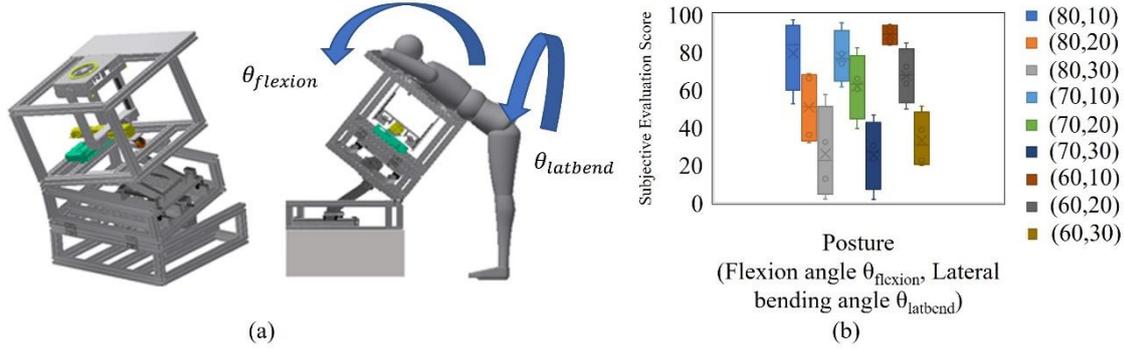

**Fig 6 (a)** Posture support mechanism for verification of body load in sitting test (b) Relationship between subjective assessment of physical strain and posture bending angles. First, the subject's flexion angle $\theta_{flexion}$ (60, 70, and 80°) and lateral flexion angle $\theta_{latbend}$ (10, 20, and 30°) were varied in all nine conditions. Next, six subjects were asked to make a subjective evaluation with VAS on whether they could maintain their posture sufficiently. The relationship between posture and physical load is analyzed based on the results. In the subjective evaluation, the case where the physical load was large, and it was challenging to maintain the posture was evaluated as zero, and the case where the posture could be maintained sufficiently was evaluated as 100.

The relationship between the subjective evaluation of physical burden and diagnostic posture is shown in Fig 6(b). The results suggest that $\theta_{latbend}$ is more responsible for the physical burden of the subject than $\theta_{flexion}$. First, the range of motion for lateral bending of the thoracolumbar region is smaller than for flexion. Second, the greater the lateral bending angle, the larger the angle between the ground and leg, and the more the load is applied to the foot base. Also, as the lateral flexion angle increases, the angle between the ground and the leg increases, and the load is placed on the foot base. As a result, the force to support the body is concentrated on the leg in the lateral bending direction, and the load is considered to increase.

Based on these results, we propose a posture angle control system to realize a seated echocardiography robot in the next section.

## 3. Diagnostic posture angle control system
### 3.1. Design concept

The preliminary results in Sec 2 proved the feasibility by adjusting the posture angle in the sitting position; the features observed in the supine and left lateral decubitus positions can be depicted. Meanwhile, it was found that the visibility variation occurred due to individual differences, and some posture angles caused the physical load on the patient. To solve those issues, we developed a posture angle control system with a seated echocardiography robot system.

The developed system consists of three active degrees of freedom (DOFs) and two passive DOFs. This system can be controlled at any posture angle while reducing the physical load on the patient (Fig 7). As described in Sec. 1, this system is available in a seated position. In addition, by using a linear motion mechanism with trapezoidal screws for all degrees of freedom, the robot maintains its posture even if the

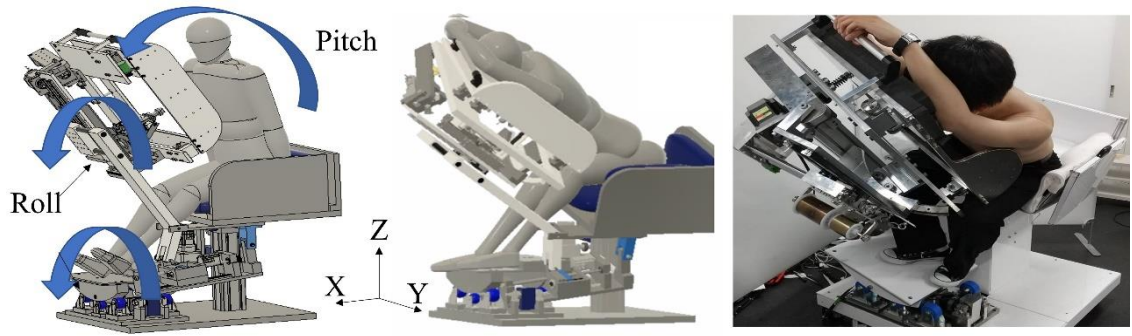

**Fig 7** Seated posture angle control system

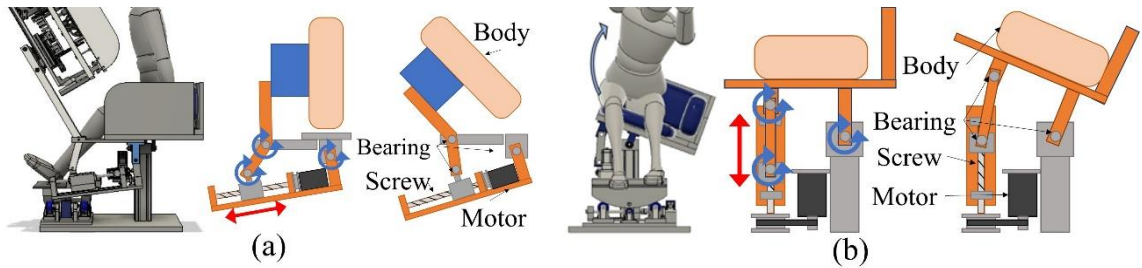

**Fig 8** (a) Lumbar bending mechanism (b) Lumbar lateral bending mechanism motor stops, thus ensuring patient safety in case of emergency. With these innovations, this system ensures safety and emergency evacuation. The pitch angle of the posture is controlled by a lumbar bending mechanism using a four-section linkage mechanism, a trapezoidal screw thread (MTSTRW16-475-F38-V12-S60-Q12, MISUMI, Japan), and a stepper motor (RKS5913RAD2-3, ORIENTAL MOTOR, Japan) (Fig 8(a)). The roll angle of the posture is controlled by two mechanisms: a lumbar lateral bending mechanism (Fig 8(b)) using a trapezoidal screw thread (MTSBRV16-232-F38-V12-S70-Q12-C30-J0, MISUMI, Japan) and a stepper motor (RKS5913RAD2-3, ORIENTAL MOTOR, Japan), and a thoracic rotation mechanism (Fig 9) using a trapezoidal screw thread (MTSBRV16-322-F32-V10-S65- Q12-C60-J3, MISUMI, Japan) and a stepper motor (RKS599RAD2-2, ORIENTAL MOTOR, Japan). In the preliminary experiment of Sec.2-1, the subjects were six males in their twenties, and the variation in the experimental conditions was limited. Considering the differences in age and gender, the differences in diagnosable posture angles due to individual differences could be even more considerable. Therefore, the required angle in this system is $\theta_{roll}$ = 0°–65° and $\theta_{pitch}$ = 0°–85°, which are larger than the diagnosable posture angles introduced in Sec 2. The thoracic rotation mechanism can be manually moved in the direction of the blue arrow shown in Fig 9 by self-weight compensation using constant load springs (CR-16.CR-19). This allows the user to adjust the height to suit their height. Additionally, to reduce the physical burden during the examination, the system is equipped with two mechanisms (1) a leg pendulum base mechanism to reduce the burden on the legs when the lateral bending angle is increased, and 2) a roll angle division by a lumbar lateral bending mechanism and a thoracic rotation mechanism. Note that red and blue arrows indicate each figure's active and passive motion parts.

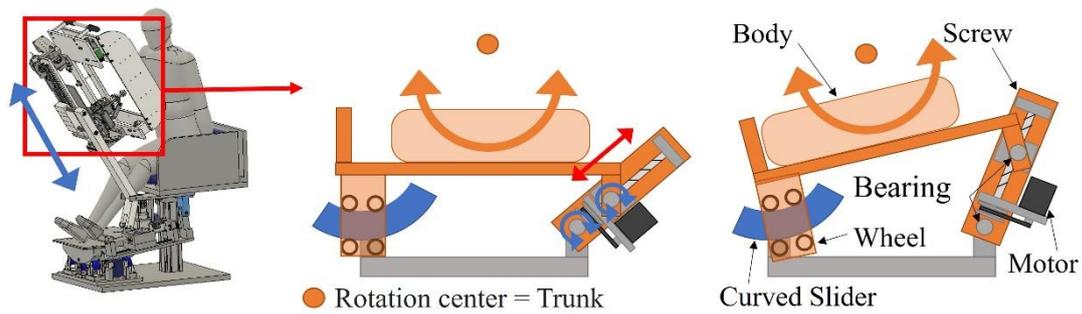

**Fig 9** Thoracic rotation mechanism

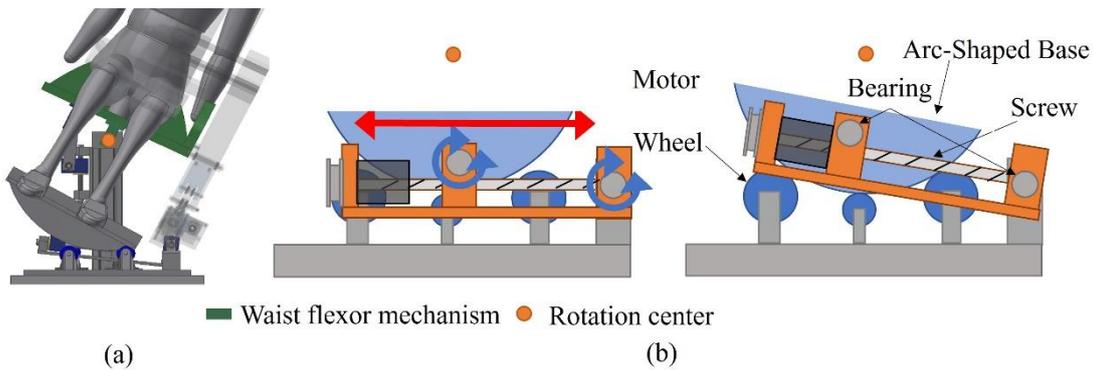

**Fig 10** Leg pendulum base mechanism; (a) Mechanism operation; (b) Mechanism details

### 3.2. Leg pendulum base mechanism

      This mechanism can change its angle in synchronization with the lateral bending angle of the lumbar lateral bending mechanism described in Sec 3-3 to maintain the legs and base vertical. As shown in Fig 10(b), the base, which has an arc at the bottom, is supported by three wheels, and the base rotates smoothly according to the side-bending angle. The base moves by a linear motion mechanism using a trapezoidal screw thread (MTSTLW16-307-F30-V12-S51-Q12, MISUMI, Japan) and a stepper motor (RKS566RAD2-3, ORIENTAL MOTOR, Japan). As shown in Fig 10(a), the center of rotation is shifted above the base, and the robot can move like a pendulum. By making the rotation centers of the waist lateral bending and this mechanism identical, the distance between the two mechanisms can be kept constant even when the lateral bending angle is increased while the waist/leg installation parts of the two mechanisms remain parallel. This allows the user to maintain a stable posture without having the legs lift off the base or having a mid-back posture when the lateral bending angle changes.

### 3.3. Lumbar lateral bending mechanism and thoracic rotation mechanism

      The leg pendulum base mechanism can eliminate the cause of the body load described in Sec. 2-2. Alternatively, a force must be applied to the legs to withstand the force of the gravity component applied in the inclined direction of the base, resulting in a foot load. The roll angle was realized by combining the lumbar lateral bending and thoracic rotation mechanisms to solve this problem. This reduces the tilt angle of the lumbar lateral bending mechanism/leg pendulum base mechanism to minimize the body load.

Moreover, the lumbar lateral bending and thoracic rotation mechanisms have been devised to enable lateral bending and rotation with minimal body load, respectively. The former mechanism is shown in Sec. 3-2. For the latter, the center of rotation is adjusted to the trunk axis of the upper body. This makes it possible to align the body's rotation with the rotation of the mechanism, which is achieved by connecting a mechanism that can move smoothly on a curved slider with four wheels and a linear motion mechanism with trapezoidal screw threads, as shown in Fig 9. As a result, the connection can be tilted when the linear motion mechanism is moved while the center of rotation is adjusted to the trunk axis.

**Table 1** List of conditions for verification of body load reduction mechanism

| Condition | Leg pendulum base mechanism | Roll ° | |
|---|---|---|---|
| | | Lumbar lateral bending mechanism | Thoracic rotation mechanism |
| A | Not introduced | 20 | 0 |
| B | Introduced | 20 | 0 |
| C | Introduced | 0 | 20 |
| D | Introduced | 10 | 10 |

## 4. Evaluation

### 4.1. Experimental setup

In this experiment, we quantitatively evaluate the physical load on each muscle when using the proposed posture angle control system and verify the efficacy of reducing the physical load by applying the leg pendulum base mechanism and dividing the roll angle using the lumbar lateral bending and thoracic rotation mechanisms. The quantitative evaluation of the body load is performed using a muscle potential measurement device (Wireless EMG system Trigno, 4 Assist, Japan). The number of subjects was 6 (height: 172 ± 16 cm; weight: 62 ± 9 kg). The detailed flow of the experiment is described below.

(1) The EMG analyzer is attached to the gastrocnemius and oblique abdominal muscles on each side.

(2) In the four conditions listed in Table 1, the subject's posture is kept at a $\theta_{g\_roll}$ of 20° and $\theta_{g\_pitch}$ of 45° for two minutes. The EMG in each part of the body at that time is recorded. Note that $\theta_{g\_roll}$ is 90° minus the angle between the gravity vector and the frontal axis, and $\theta_{g\_pitch}$ is 90 - the angle between the gravity vector and the sagittal axis.

(3) A band-pass filter with a cutoff frequency of 20 Hz and 450 Hz is applied to the EMGs acquired under each condition, and the signal envelope using the root mean square value of a 300 ms moving slit is extracted. Note that this process was performed with reference to Ref. [15].

(4) The median of the values calculated in (3) is computed, and this value is used as the measured EMG in each condition. Next, the sum of the measured EMG of the left and right gastrocnemius and oblique abdominal muscles in each condition is calculated as the leg and abdominal loads, respectively. Then, Conditions A and B in Table 1 are compared to verify the effectiveness of the leg pendulum base mechanism; Conditions B, C, and D are compared to verify the effectiveness of reducing the physical load by dividing the roll angle by the lumbar lateral flexion and thoracic rotation mechanisms; and Conditions A and D are compared to verify the effectiveness of combining the two physical load reduction mechanisms.

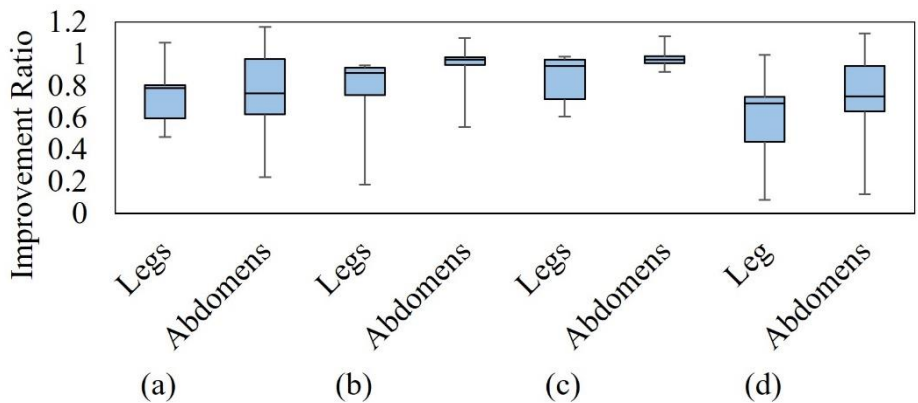

**Fig 11** Percentage of leg/abdominal load for each condition; (a) Ratio of Condition B to Condition A; (b) Ratio of Condition D to Condition B; (c) Ratio of Condition D to Condition C; (d) Ratio of Condition D to Condition A

Note that the low EMG value demonstrates the low physical load.

### 4.2. Results

The leg/abdomen load ratio in Conditions B and D to Condition A is shown in Figs 14(a) and (d), and the ratio of the leg/abdomen load in Condition D to that in Conditions B and C is shown in Figs 14(b) and (c), respectively. The median ratio of leg/abdomen load for Condition B to Condition A (Fig 11 (a)) was 0.785 for the legs and 0.754 for the abdomen. Additionally, the median ratio of leg/abdomen load for Condition D to Condition B (Fig 11 (b)) was 0.879 for the legs and 0.965 for the abdomen. The median ratio of leg/abdomen load for Condition D to Condition C (Fig 11 (c)) was 0.924 for the legs and 0.964 for the abdomen. Lastly, the median ratio of leg/abdomen load for Condition D to Condition A (Fig 11 (d)) was 0.691 for the legs and 0.736 for the abdomen.

### 4.3. Discussion

The results of Sec. 4-2 suggested that the leg pendulum base, lumbar lateral bending, and thoracic rotation mechanisms effectively reduced the body load by dividing the roll angle. On the other hand, it was found that some subjects disagreed with the effectiveness of these mechanisms in reducing body load due to individual differences in the muscles of the human body. For example, if the leg strength is superior to the abdominal muscles, the load may be less in Condition B than in Condition D. Therefore, it would be possible to reduce the load on the body by providing an appropriate ratio of the load on the trunk, leg strength, and abdominal muscles for each individual.

The limitations of this study are discussed following. First, the physical loads were only evaluated for the short period and needed to be assessed over a longer period. We will investigate the relationship between the physical load measured with EMGs and the difficulty of maintaining the posture angle for a long time. The second limitation is that the posture angle control system was developed based on the limited number of subjects. There may be subjects who cannot be examined in the sitting posture due to individual differences. We recognize it is necessary to perform a comparative study on a more significant number of subjects with variations in gender, body size, and age. The third limitation is that the effects on the obtained image due to the change of the examination position needs to be more investigated. Although it has been confirmed that the necessary diagnostic features can be observed in the sitting posture, there might be some other effects. For example, the position and expanding way of the lung may be changed due to the examination posture, which increases the degree to which the lungs cover the heart and makes it difficult to obtain a clear image. In some cases, the change of the examination posture may alter the direction of the load on the heart due to gravity, which compresses the heart and affects the pulsation.

## 5. Conclusion

This manuscript presents a feasibility study of seated-style echocardiography. Conventional robotic ultrasound systems were utilized with patients in supine positions since their configurations are mostly serial robotic arm and gantry-style. Meanwhile, the limitation of the systems is that it is difficult to evacuate the patients in case of emergency (e.g., patient discomfort and system failure) because the patients are restricted between the robot system and bed. Then, it is ideal that the patient undergoes the examination in the sitting position in terms of safety. Preliminary results showed that adjusting the diagnostic posture angle enabled us to obtain the views, including cardiac disease features, as in the conventional examination. Also, the results showed that a physical burden occurred for patients depending on their posture angle. Based on those results, we proposed a seated posture control system to adjust the diagnostic posture without causing any physical load during the examination. Two unique mechanisms to reduce the physical load were incorporated into the system: (1) a leg pendulum base mechanism to reduce the load on the legs when the lateral bending angle increases, and (2) a roll angle division by a lumbar lateral bending mechanism and a thoracic rotation mechanism. Experimental results demonstrated that those mechanisms could reduce the body load, which occurred in the seated echocardiography, and showed the potential of the seated-style echocardiography robot. In the future, we plan to apply the posture angle control system and the ultrasound probe scanning mechanism to automatic echocardiography using robots.

## Acknowledgment

This research is supported by the NSK Foundation for Advancement of Mechatronics and Institute for Mechanical Engineering Frontiers.

## References

1. WHO (World Health Organization) (2021) Cardiovascular diseases (CVDs).
2. Aly I, Rizvi A, Roberts W, Khalid S, Kassem MW, Salandy S, Plessis M, Tubbs RS, Loukas M (2021) Cardiac ultrasound: An Anatomical and Clinical Review, Translational Research in Anatomy. vol. 22. https://doi.org/10.1016/j.tria.2020.100083
3. Koizumi N, Warisawa S, Nagoshi M, Hashizume H, Mitsuishi M (2009) Construction methodology for a remote ultrasound diagnostic system. IEEE Trans. Robot., vol. 25, no. 3, pp. 522–538. 10.1109/TRO.2009.2019785
4. Abolmaesumi P, Salcudean SE, Zhu WH, Sirouspour MR, DiMaio SP (2002) Image-guided control

of a robot for medical ultrasound. IEEE Trans. Robot. Autom., vol. 18, no. 1, pp. 11–23. 10.1109/70.988970

5. Ito K, Sugano S, Iwata H: Portable and attachable tele-echography robot system (2010) FASTele. 2010 Annu. Int. Conf. IEEE Eng. Med. Biol. Soc. EMBC'10, pp. 487–490. 10.1109/IEMBS.2010.5627123

6. Mustafa ASB, Ishii T, Matsunaga Y, Nakadate R, Ishii H, Ogawa K, Saito A, Sugawara M, Niki K, an Takanishi A (2013) Development of robotic system for autonomous liver screening using ultrasound scanning device. IEEE Int. Conf. Robot. Biomimetics, pp. 804–809. 10.1109/ROBIO.2013.6739561

7. Fang TY, Zhang HK, Finocchi R, Taylor RH, Boctor EM (2017) Force-assisted ultrasound imaging system through dual force sensing and admittance robot control. Int. J. Comput. Assist. Radiol. Surg., vol. 12, no. 6, pp. 983–991. 10.1007/s11548-017-1566-9

8. Arbeille P, Ruiz J, Herve P, Chevillot M, Poisson G, Perrotin F (2005) Fetal tele-echography using a robotic arm and a satellite link. Ultrasound Obstet. Gynecol., vol. 26, no. 3, pp. 221–226. https://doi.org/10.1002/uog.1987

9. Esteban J, Simson W, Witzig SR, Rienmüller A, Virga S, Frisch B, Zettinig O, Sakara S, Ryang Y, Navab N and Hennersperger C (2018) Robotic ultrasound-guided facet joint insertion. Int. J. Comput. Assist. Radiol. Surg.,vol. 13, no. 6, pp. 895–904. 10.1007/s11548-018-1759-x

10. Wang S, Singh D, Johnson D, Althoefer K: Robotic Ultrasound (2016) View Planning, Tracking, and Automatic Acquisition of Transesophageal Echocardiography. IEEE Robot. Autom. Mag., vol. 23, no. 4, pp. 118–127. 10.1109/MRA.2016.2580478

11. Wang S, Housden J, Noh Y, Singh D, Singh A, Skelton E, Matthew J, Tan C, Back J, Lindenroth L, Gomez A (2019) Robotic-assisted ultrasound for fetal imaging: Evolution from single-arm to dual-arm system. Lect. Notes Comput. Sci. (including Subser. Lect. Notes Artif. Intell. Lect. Notes Bioinformatics), vol. 11650 LNAI, no. February, pp. 27–38. https://doi.org/10.1007/978-3-030-25332-5_3

12. Wang S, Housden RJ, Noh Y, Singh A, Lindenroth L, Liu H, Althoefer K, Hajnal J, Singh D, Rhode K (2019) Analysis of a customized clutch joint designed for the safety management of an ultrasound robot. Appl. Sci., vol. 9, no. 9, pp. 1–16. https://doi.org/10.3390/app9091900

13. Takachi Y, Masuda K, Yoshinaga T, Aoki Y (2011) Development of a Support System for Handling Ultrasound Proble to Alleviate Fatigue of Physician by Introducing a Coordinated Motio with robot. Journal of the Robotics Society of Japan, vol. 29, no. 7, pp.634-642.

14. Vegas A (2018) Perioperative Two-Dimensional Transesophageal Echocardiography: A Practical Handbook. Springer.

15. Moreira L, Figueiredo J, Fonseca P, Vilas-Boas JP, Santos CP (2021) Lower limb kinematic, kinetic, and EMG data from young healthy humans during walking at controlled speeds. Scientific Data, vol. 8, Article number 103. https://doi.org/10.6084/m9.figshare.13169348